\documentclass{article}

\usepackage{PRIMEarxiv}

\usepackage[utf8]{inputenc} 
\usepackage[T1]{fontenc}    
\usepackage{hyperref}       
\usepackage{url}            
\usepackage{booktabs}       
\usepackage{amsfonts}       
\usepackage{nicefrac}       
\usepackage{microtype}      
\usepackage{lipsum}
\usepackage{fancyhdr}       
\usepackage{graphicx}       
\usepackage{amsmath}
\usepackage{xcolor}
\usepackage{algorithm, algorithmic}
\usepackage{subcaption}
\usepackage{amssymb}
\usepackage{tablefootnote}
\usepackage{multirow}

\graphicspath{{media/}}     

\title{CoDefend: Cross-Modal Collaborative Defense via Diffusion Purification and Prompt Optimization
}

\author{%
  Fengling Zhu \\
  School of Computer Science\\
  Nanjing University\\
  Nanjing, Jiangsu 210023 \\
  \texttt{fenglingzhu@smail.nju.edu.cn} \\
  \And
  Boshi Liu \\
  Nanjing University \\
  Nanjing, Jiangsu 210023 \\
  \texttt{liubs@smail.nju.edu.cn} \\
  \AND
  Jingyu Hua \\
  Nanjing University \\
  Nanjing, Jiangsu 210023 \\
  \texttt{huajingyu@nju.edu.cn} \\
  \And
  Sheng Zhong \\
  Nanjing University \\
  Nanjing, Jiangsu 210023 \\
  \texttt{zhongsheng@nju.edu.cn} \\
}

\begin{document}
\maketitle

\begin{abstract}
  Multimodal Large Language Models (MLLMs) have achieved remarkable success in tasks such as image captioning, visual question answering, and cross-modal reasoning by integrating visual and textual modalities. However, their multimodal nature also exposes them to adversarial threats, where attackers can perturb either modality—or both jointly—to induce harmful, misleading, or policy-violating outputs. Existing defense strategies, such as adversarial training and input purification, face notable limitations: adversarial training typically improves robustness only against known attacks while incurring high computational costs, whereas conventional purification approaches often suffer from degraded image quality and insufficient generalization to complex multimodal tasks.

In this work, we focus on defending the visual modality, which frequently serves as the primary entry point for adversarial manipulation. We propose a supervised diffusion-based denoising framework that leverages paired adversarial–clean image datasets to fine-tune diffusion models with directional, task-specific guidance. Unlike prior unsupervised purification methods such as DiffPure, our approach achieves higher-quality reconstructions while significantly improving defense robustness in multimodal tasks. Furthermore, we incorporate prompt optimization as a complementary defense mechanism, enhancing resistance against diverse and unseen attack strategies.

Extensive experiments on image captioning and visual question answering demonstrate that our method not only substantially improves robustness but also exhibits strong transferability to unknown adversarial attacks. These results highlight the effectiveness of supervised diffusion-based denoising for multimodal defense, paving the way for more reliable and secure deployment of MLLMs in real-world applications.

\end{abstract}


\section{Introduction}
In recent years, Multimodal Large Language Models (MLLMs) have achieved remarkable breakthroughs in the field of artificial intelligence. By integrating visual and linguistic modalities, these models demonstrate strong cross-modal understanding and generation capabilities, and have been widely applied in tasks such as visual question answering, image captioning, and visual reasoning \cite{lu2019vilbert, li2019visualbert, su2019vl}. Representative examples include BLIP \cite{li2022blip}, BLIP-2 \cite{li2023blip2}, MiniGPT-4 \cite{zhu2023minigpt}, GPT-4V \cite{yan2023gpt}, and the LLaVA series \cite{liu2023llava}, which generally adopt the “image encoder + large language model + cross-modal alignment” architecture to achieve unified semantic modeling and natural language generation across multimodal inputs.

However, while MLLMs push the boundaries of model capabilities, they also introduce more complex and subtle security risks. Compared with traditional language models, the attack surface of MLLMs is significantly expanded. Attackers can not only manipulate text inputs, but also inject perturbations into the image modality, or even craft joint image-text perturbations to bypass content moderation mechanisms or manipulate model outputs \cite{dong2023robustgooglesbardadversarial,luo2024anCroPA,yin2023vlattack,zhang2022towards}. Thus, enhancing the robustness of multimodal large models against adversarial inputs and developing efficient, generalizable defense mechanisms has become a core challenge in current multimodal AI research.

Previous studies have proposed a variety of defense strategies for multimodal models. For example, adversarial training introduces perturbed samples during training, enabling the model to learn resistance against known attacks \cite{goodfellow2014explaining, kannan2018adversarial,madry2019deeplearningmodelsresistant,wang2025boosting}. However, such approaches face two key limitations: (1) adversarial training often improves robustness only against specific or known attacks, with limited effectiveness against unknown or out-of-distribution perturbations; (2) the approach typically requires generating and iteratively incorporating large volumes of adversarial samples during training, leading to heavy computational costs that hinder large-scale deployment in resource-constrained settings. Therefore, how to achieve generalizable defenses against multimodal adversarial attacks while maintaining efficiency remains an open problem.

In multimodal attacks, the image modality is often the most vulnerable entry point for adversaries, making image purification an increasingly important and effective defense strategy. The core idea is to preprocess input images before model inference, thereby weakening or removing potential adversarial perturbations and reducing attack success rates. Compared with adversarial training or other model-dependent defenses, image purification is model-agnostic and computationally lightweight, making it highly suitable for direct deployment at inference time. As such, it demonstrates greater flexibility and generality in practical applications.

Existing image purification approaches can be broadly divided into three categories. The first category is signal-processing-based strategies, such as JPEG compression, filtering, downsampling, and resampling, which improve robustness by disrupting the fine-grained patterns that perturbations rely on \cite{xie2018mitigating}. The second category is learning-based approaches, which train specialized denoising networks or detection–correction modules to identify and remove perturbations, achieving good performance under specific attack settings. The third category leverages image transformation and denoising through generative or diffusion models. In particular, image reconstruction via diffusion models has recently gained significant attention, as it purifies inputs at the feature level and restores representations closer to the original semantics \cite{nie2022diffpure,pmlr-v235-bai24b,zhao2025bluesuffix,10.1145/3731715.3733367}, offering unique advantages against complex perturbations.

Among existing studies, DiffPure \cite{nie2022diffpure} is one of the most representative diffusion-based denoising approaches. It injects small Gaussian noise into adversarial samples and employs the reverse diffusion process to gradually reconstruct the image, thereby removing adversarial perturbations while retaining semantic information. However, subsequent theoretical analysis revealed inherent limitations: the divergence between purified data and clean data can exceed that of the original adversarial perturbations \cite{pmlr-v235-bai24b}, leading to low-quality reconstructions. To address this issue, \cite{pmlr-v235-bai24b} introduced a contrastive learning-based refinement, guiding the diffusion model to perform more effective denoising and significantly lowering attack success rates in image classification tasks. Nevertheless, such methods remain primarily focused on classification, with limited defensive effectiveness for more complex vision-language tasks such as visual question answering.

To overcome these limitations, this paper proposes a supervised diffusion-based denoising approach. Specifically, we construct a paired dataset of adversarial–clean images and fine-tune Stable Diffusion on this dataset, enabling the model to learn directional and task-specific denoising capabilities during training. Unlike traditional unsupervised diffusion denoising, the proposed method not only achieves superior reconstruction quality, but also substantially enhances robustness in multimodal tasks, particularly in image captioning and visual question answering, where it exhibits stronger generalizability and defense performance.

Experiments on image captioning and visual question answering tasks demonstrate that our method not only improves robustness effectively but also shows strong transferability, resisting a variety of unseen attack strategies. The main contributions of this paper are summarized as follows:

We propose a supervised, high-quality image adversarial denoiser and apply it to visual question answering. Compared with classification, visual question answering involves more complex multimodal interactions, making robustness improvements both more challenging and more impactful.

In addition to the core denoising method, we introduce prompt optimization as a complementary enhancement to further improve defensive effectiveness.

We conduct systematic evaluations on image captioning and visual question answering, and the experimental results verify the effectiveness and superiority of the proposed method in improving robustness and defensive transferability.

\section{Related Work}
This section reviews adversarial attacks and defense mechanisms for multimodal large language models (VLMs), with a particular focus on common attack strategies and existing defense methods.

Multimodal Adversarial Attacks. With the wide deployment of large-scale vision-language models (e.g., GPT-4V, Gemini, LLaVA), their multimodal input capabilities significantly enhance task performance but also introduce more complex attack surfaces. Moreover, the objectives of adversarial attacks have expanded from simple classification to more sophisticated text generation and cross-modal reasoning tasks. Studies have shown that adversaries can manipulate image and text modalities separately or jointly to induce models into producing incorrect, harmful, or even prohibited outputs. For instance, Dong et al. \cite{dong2023robustgooglesbardadversarial} introduced fine-grained image perturbations to force the model to generate malicious content, achieving targeted attacks. VLAttack \cite{yin2023vlattack} further demonstrated the effectiveness of joint perturbations on both image and text modalities, making models more susceptible to misleading outputs. CroPA \cite{luo2024anCroPA} incorporated prompt learning into perturbation generation, enabling generalized misinterpretations even for unseen prompt combinations. Collectively, these works reveal that the joint manipulation of image and text modalities can greatly amplify attack effectiveness and transferability.

In summary, existing research highlights the synergistic impact of image-text perturbations in multimodal adversarial attacks. However, among these, the image modality often plays a particularly critical role: even subtle visual perturbations can substantially alter model reasoning and generation, amplifying the overall impact of cross-modal attacks. Consequently, researchers have initially focused on defenses for the visual modality, seeking to mitigate adversarial effects and improve robustness through techniques such as image preprocessing, robust feature extraction, and input denoising.

Given that the image modality is the most exploitable entry point for adversaries, image purification has gradually emerged as an important and effective defense strategy. The core idea is to preprocess input images before inference to suppress or remove potential adversarial perturbations, thereby reducing attack success rates. Compared with adversarial training and other model-dependent approaches, image purification is model-agnostic and incurs relatively low computational overhead at inference time, making it well-suited for deployment in practical systems with higher flexibility and generality.

Existing research generally categorizes image purification methods into three classes:

Signal processing-based approaches, such as JPEG compression, smoothing filters, downsampling, and resampling, which disrupt fine-grained perturbation patterns to enhance robustness \cite{xie2018mitigating}. These methods are simple and efficient but are often less effective against adaptive attacks.

Learning-based approaches, which train dedicated denoising networks or detection–restoration modules to automatically identify and remove perturbations. While effective in specific attack scenarios, their generalization ability depends heavily on training data distributions, leading to limited performance against unseen attacks.

Generative or diffusion-based approaches, which have gained increasing attention in recent years. These methods leverage image reconstruction to “purify” inputs at the feature level, recovering representations closer to the original semantics \cite{nie2022diffpure,pmlr-v235-bai24b,zhao2025bluesuffix,10.1145/3731715.3733367}.

Among diffusion-based approaches, DiffPure \cite{nie2022diffpure} is the most representative. It injects small Gaussian noise into adversarial samples and applies the reverse diffusion process to gradually reconstruct images, thereby removing adversarial perturbations while retaining semantic information. However, subsequent theoretical analyses revealed inherent limitations: the upper bound of differences between purified and clean samples may even exceed that of the original adversarial perturbation \cite{pmlr-v235-bai24b}, leading to suboptimal image quality. To address this, \cite{pmlr-v235-bai24b} proposed a contrastive learning enhancement, guiding the diffusion model to perform more effective denoising, significantly reducing attack success rates in classification tasks. Nevertheless, these improvements remain primarily focused on classification scenarios, with limited effectiveness in more complex vision-language tasks such as visual question answering and cross-modal reasoning. Similarly, multimodal defense frameworks like BlueSuffix \cite{zhao2025bluesuffix} incorporate DiffPure directly in their visual defense stage, inheriting its drawbacks and resulting in subpar purified image quality.

To overcome these limitations, this paper proposes a supervised diffusion-based denoising method. Specifically, we construct paired datasets of adversarial samples and their corresponding clean images, and fine-tune Stable Diffusion to learn task-oriented and directional denoising during training. Unlike traditional unsupervised diffusion-based purification, our approach achieves superior image restoration quality while significantly enhancing defense performance in multimodal tasks. In particular, it demonstrates stronger robustness and generalizability in representative applications such as image captioning and visual question answering.

\section{Methodology}
\subsection{Preliminaries}
\textbf{Threat Model.} Let \( G \) represent a multimodal large language model (MLLM), which takes an image \( \mathbf{I} \) and a text prompt \( \mathbf{T} \) as inputs and generates a natural language response. An attacker can inject small, imperceptible perturbations into \( \mathbf{I} \) or \( \mathbf{T} \) to manipulate the model’s output, either weakening performance (usability attacks) or inducing undesirable content (jailbreak attacks).

Let \( \mathbf{y} \) denote the model's response to normal inputs, and \( \mathbf{y'} \) the response to perturbed inputs \( \mathbf{I'} \) and \( \mathbf{T'} \), where \( \mathbf{I'} = \mathbf{I} + \delta_I \) and \( \mathbf{T'} = \mathbf{T} + \delta_T \). The goal is to classify attacks as either untargeted or targeted.

In untargeted attacks, the goal is to cause a deviation in the model’s output, i.e., \( \mathbf{y'} \neq \mathbf{y} \), formalized as:
\begin{equation}
    \min_{\delta_I, \delta_T} \mathcal{L}\left(G\left(\mathbf{I} + \delta_I, \mathbf{T} + \delta_T\right), \mathbf{y}\right) \quad \text{s.t.} \quad \|\delta_I\| \leq \epsilon_I, \|\delta_T\| \leq \epsilon_T
\end{equation}
For targeted attacks, the aim is to force the model to generate a specific target response \( \hat{\mathbf{y}} \), i.e., \( \mathbf{y'} = G(\mathbf{I'}, \mathbf{T'}) = \hat{\mathbf{y}} \), commonly seen in jailbreak attacks, where the attacker aims to bypass the model's filtering mechanism:
\begin{equation}
    \min_{\delta_I, \delta_T} \mathcal{L}\left(G\left(\mathbf{I} + \delta_I, \mathbf{T} + \delta_T\right), \hat{\mathbf{y}}\right) \quad \text{s.t.} \quad \|\delta_I\| \leq \epsilon_I, \|\delta_T\| \leq \epsilon_T
\end{equation}

In this study, we focus on black-box adversarial defense, assuming the defender does not have access to the model's parameters, gradients, or structural information, and can only obtain outputs corresponding to inputs through limited queries. In this setup, the attacker may use either white-box or black-box methods to generate adversarial samples in the image or text modalities, thereby misleading the multimodal large language model into producing erroneous or inappropriate responses.

\textbf{InstructPix2Pix}. The original diffusion models (e.g., DDPM \cite{ho2020denoisingdiffusionprobabilisticmodels}) model high-resolution images in pixel space, and while they have strong generative capabilities, they incur significant computational costs, leading to low training and inference efficiency. To alleviate this issue, Latent Diffusion Models (LDMs) \cite{9878449LDMs} proposed transferring the diffusion process from the high-dimensional pixel space to a more compact latent space, significantly reducing computational costs while maintaining the quality of the generated images. The core idea of this method is to use a pre-trained variational autoencoder (VAE) \cite{kingma2022autoencodingvariationalbayes} to compress the original image into a latent representation and then perform diffusion modeling in the latent space.

Specifically, the autoencoder consists of an encoder $\mathcal{E}$ and a decoder $\mathcal{D}$. The encoder maps the input image $x$ to a latent vector $z$, i.e., $z = \mathcal{E}(x)$, and the decoder is responsible for reconstructing the image from the latent vector, i.e., $\hat{x} = \mathcal{D}(z)$. This latent space modeling mechanism forms the basis for current mainstream generative models such as Stable Diffusion\cite{rombach2022high}.

Within this framework, InstructPix2Pix \cite{InstructPix2Pix10204579} was the first to propose fine-tuning Stable Diffusion to enable the model to perform semantic-level editing on input images based on natural language instructions. The training objective is to learn a conditional noise prediction network $\varepsilon_\theta$ to predict additive noise in the latent space, with the prediction process depending on both the image condition $c_I$ and the text instruction condition $c_T$. The training objective is optimized as follows:
\begin{equation}
L=\mathbb{E}_{\mathcal{E}(x),\mathcal{E}(c_{I}),c_{T},\epsilon\sim\mathcal{N}(0,1),t}\left[\|\epsilon-\epsilon_{\theta}(z_{t},t,\mathcal{E}(c_{I}),c_{T}))\|_{2}^{2}\right]
\end{equation}
where $z_t$ is the noisy latent vector at time step $t$, and $\epsilon$ is the true Gaussian noise, with the model learning the reverse diffusion process by fitting this noise.

InstructPix2Pix simplifies image editing by requiring only an image \$c\_I\$ and a text instruction \$c\_T\$, generating results in a single pass without the need for masks, target images, or fine-tuning. This makes it efficient and ideal for human-computer interaction.However, the pre-trained model excels at general tasks like style transfer and semantic replacement but struggles with removing adversarial perturbations, particularly fine-grained noise, due to limited robustness and denoising capabilities.

To address this, we further fine-tunes InstructPix2Pix, enabling it not only to understand "remove adversarial noise" instructions but also to effectively recognize and remove adversarial perturbations from images in latent space, while preserving the original semantic structure and usability of the images. This design improves the model's stability and practicality when handling adversarial samples and provides an efficient image purification solution for enhancing the security of multimodal models.
\begin{figure}[t]\centering
    \centering
    \includegraphics[width=\linewidth]{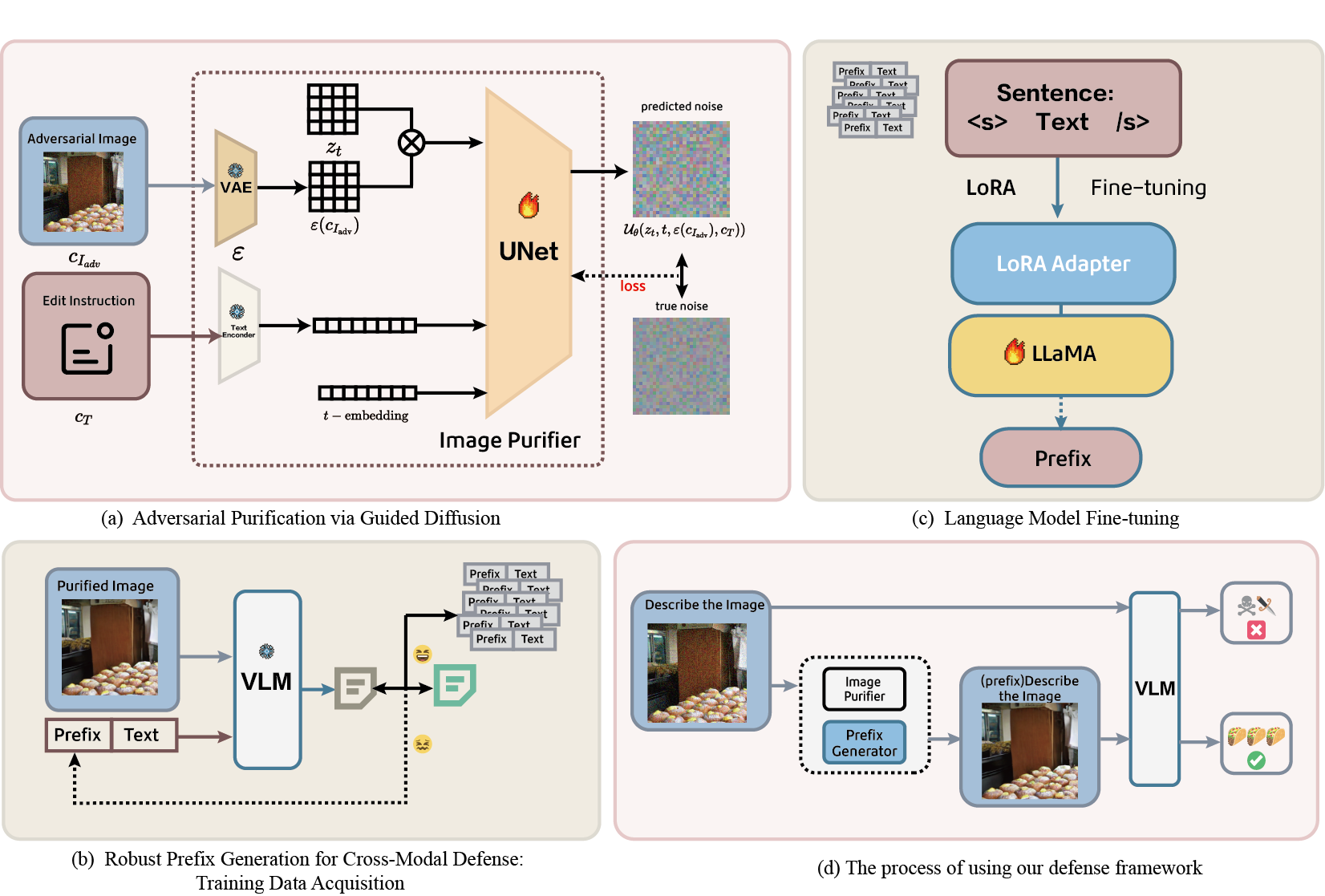}
    \caption{An overview of CoDefend.}
    \label{fig:Overview}
\vspace{-1.5em}
\end{figure}
\subsection{Adversarial Purification via Guided Diffusion}
\label{subsec:diffusion}
\begin{figure}[t]\centering
\begin{subfigure}[t]{0.49\textwidth}
\centering
\includegraphics[width=\linewidth]{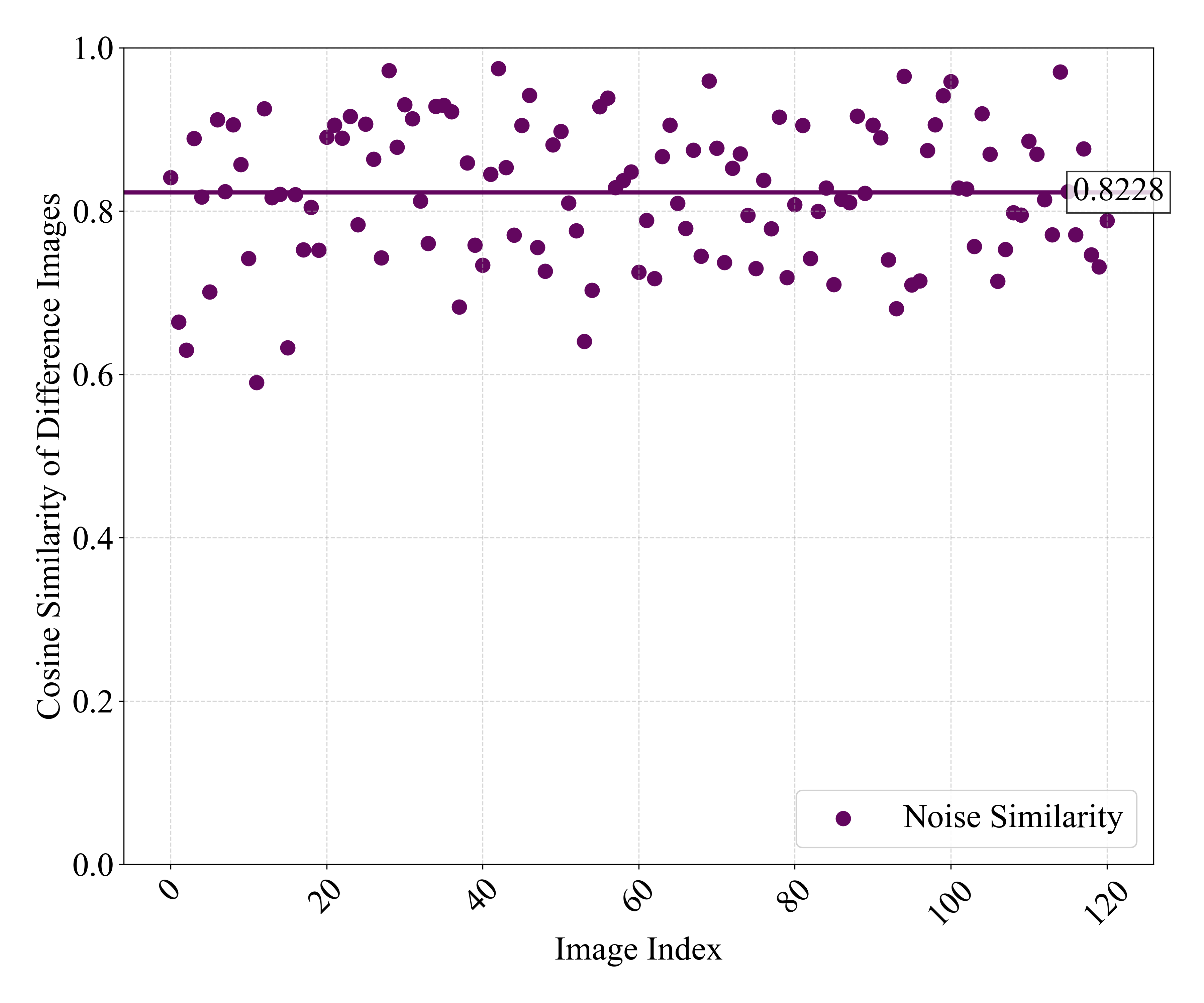}\caption{Scatter plot depicting the similarity between adversarial perturbations and removed noise.}
\label{fig:semantic_dist_c}
\end{subfigure}
\hfill
\begin{subfigure}[t]{0.49\textwidth}
\centering
\includegraphics[width=\linewidth]{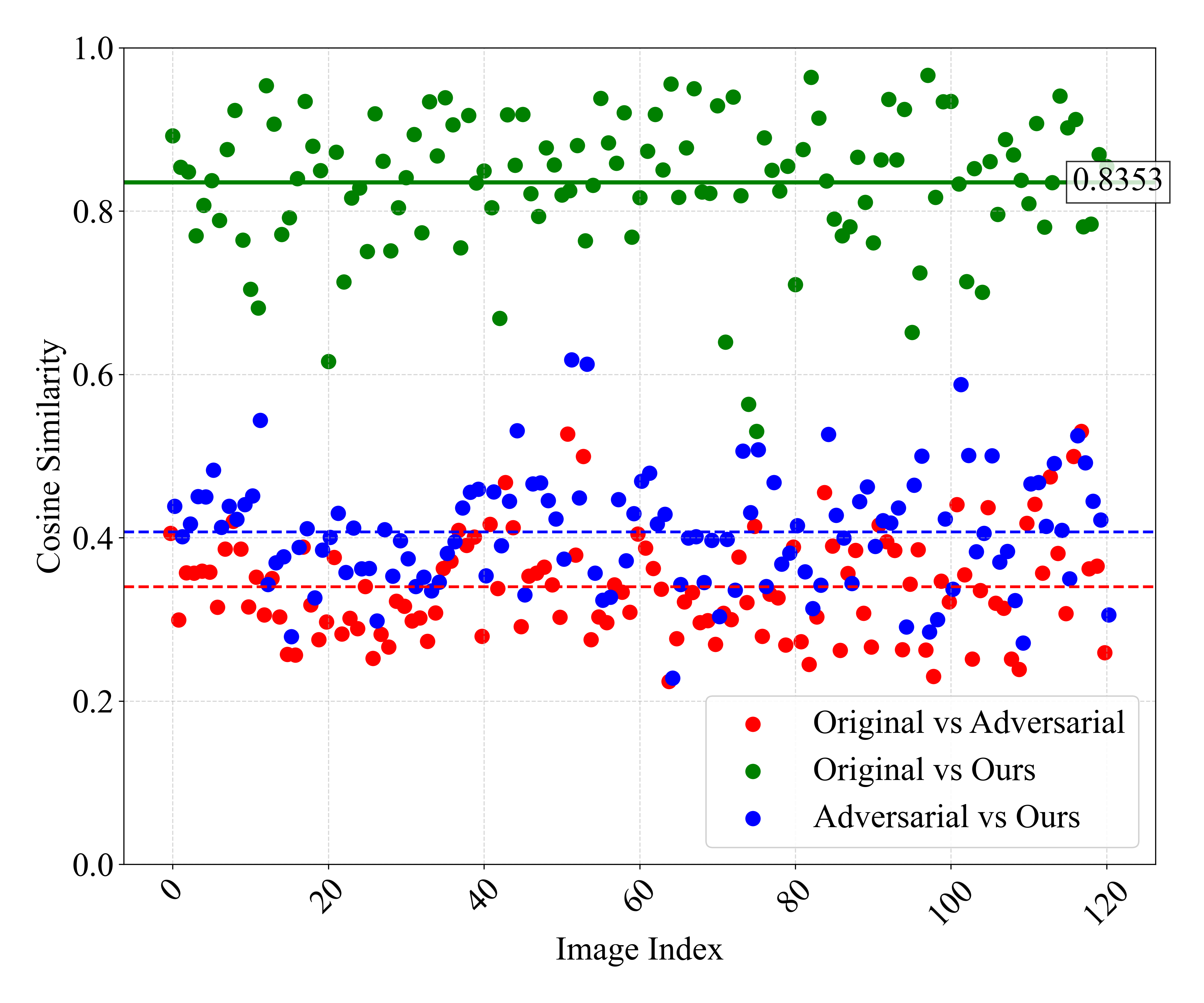}\caption{Scatter plot depicting the similarity among original, adversarial, and purified images using our technique.}
\label{fig:semantic_dist_a}
\end{subfigure}
\caption{An assessment of feature-level purification efficacy of our approach was conducted on 121 adversarial samples generated through the BARD attack \cite{dong2023robustgooglesbardadversarial}, utilizing the visual encoder of BLIP-2.}
\label{fig:noise_analysis}
\vspace{-1.5em}
\end{figure}

We develop a diffusion-based purification framework that learns to remove adversarial perturbations through conditional denoising while preserving semantic fidelity. Our approach leverages two intrinsic properties of adversarial attacks in vision-language systems: (1) their measurable deviations in joint feature-level spaces, as evidenced by the 0.340 vision-language feature similarity between adversarial samples and original images shown in Figure~\ref{fig:semantic_dist_a}, and (2) the structured patterns in perturbation distribution that enable discriminative removal.

The framework operates through a modified UNet architecture \(\mathcal{U}_\theta\) trained on paired adversarial-original images \((I_{\text{adv}}, I_{\text{clean}})\), with the objective:
\begin{equation}
\mathcal{L}_{\text{diff}} = \mathbb{E}_{\epsilon,t}\left[ \left\lVert \epsilon - \mathcal{U}_{\theta}(z_t, t, \varepsilon (c_{I_{\text{adv}}}),c_{T})) \right\rVert_2^2 \right],
\label{eq:diff_loss}
\end{equation}
where \(z_t\) contains perturbed clean latent features at diffusion step \(t\), and \(\varepsilon(c_{I_{\text{adv}}})\) encodes adversarial context, \(c_T\) representing the image editing instruction. 

To maintain semantic fidelity, we freeze all pre-trained vision-language components and restrict optimization to the denoising network parameters. This ensures stable feature representations while adapting the purification process to multimodal defense requirements.

As quantified in Figure~\ref{fig:noise_analysis}, empirical analysis confirms our method's efficacy in recovering semantic fidelity and discriminatively removing adversarial noise. This approach ensures that the model maintains its generative capacity while learning to suppress adversarial perturbations, establishing a foundation and the most important part for the multi-stage defense pipeline introduced in this work. Given that the discussion takes place in a multimodal context, we also suggest a defensive strategy as a supplementary measure for image purification within the text modality.

\subsection{Robust Prefix Generation for Cross-Modal Defense}
\label{subsec:framework}

Our defense framework centers on learning robust textual prefixes that complement visual purification. The core methodology trains a LLaMA-2 7B model \cite{touvron2023llama} to generate protective prefixes using adversarial-resistant vision-language pairs, constructed through three coordinated steps:

\textbf{Training Data Acquisition.} To enable prefix learning, we first create paired data of purified images and optimized prompts through two parallel processes. On the visual front, diffusion-based purification removes adversarial perturbations from input $x$ to produce $\hat{x}_{\text{clean}}$, preserving semantic content through stochastic differential equations. For textual enhancement, we optimize defense-aware prompts $p_{\text{opt}}$ via gradient-guided alignment with BLIP/BLIP-2 models:
\begin{equation}
p_{\text{opt}} = \arg\min_{p\in\mathcal{P}} \mathcal{L}_{\text{CE}}(\mathcal{B}(\hat{x}_{\text{clean}};p\oplus q), y_{\text{true}})
\end{equation}
where the prompt space $\mathcal{P}$ is constrained by trigger tokens and semantic consistency bounds $\lVert p_{\text{opt}} - p_{\text{base}} \rVert_{2} \leq \delta$. This joint optimization employs HotFlip-based token replacement with multi-round beam search, balancing attack resistance and linguistic naturalness.

\textbf{Language Model Fine-tuning.} The generated pairs $\{(\hat{x}^{i}_{\text{clean}},p^{i}_{\text{opt}})\}$ train a LLaMA-2 7B model to predict protective prefixes conditioned on visual-textual context. We implement parameter-efficient tuning through Low-Rank Adaptation (LoRA) on query/value projections:
\begin{equation}
\mathcal{L}_{\text{LoRA}} = \mathbb{E}_{(q,p)}\left[ -\sum_{t=1}^T \log P_{\mathcal{L}}(p_t \mid q, p_{<t}) \right]
\end{equation}
with rank $r=8$ and scaling factor $\alpha=16$. The instructional template \texttt{"Generate a protective prefix for: <query>"} explicitly conditions prefix generation on attack scenarios. During deployment, the model prepends learned prefixes $p_{\text{gen}}$ to user queries, modifying input semantics before reaching victim models.

The framework still operates in standard black-box settings: prompt optimization relies on open-source BLIP/BLIP-2. No architectural knowledge of protected models is needed during training or inference.

As outlined in Algorithm~\ref{alg:overall}, the complete framework processes adversarial images through three sequential stages: first, diffusion-based purification removes visual perturbations; second, cross-modal prompt optimization generates prefixes that semantically anchor the purified images; finally, the language model aligns these prefixes with input queries to construct robust inputs for the downstream visual-language model. This cohesive architecture ensures adversarial perturbations are mitigated across modalities, enabling reliable predictions from contaminated inputs.

\begin{algorithm}[htb]
\caption{Framework of CoDefend}
\label{alg:overall}
\begin{algorithmic}[1]
    \REQUIRE Adversarial image $x_{\text{adv}}$, input query $q$, target text $y_{\text{true}}$
    \ENSURE Robust prediction $\hat{y}$
    \STATE $\hat{x}_{\text{clean}} \leftarrow \text{DiffusionPurify}(x_{\text{adv}})$ \COMMENT{Remove adversarial noise via guided diffusion (Sec~\ref{subsec:diffusion})}
    \STATE $p_{\text{gen}} \leftarrow \text{PrefixGenerate}(q, \hat{x}_{\text{clean}})$ \COMMENT{Generate defense prefix with fine-tuned LLaMA-2 (Sec~\ref{subsec:framework})}
    \STATE $\hat{y} \leftarrow \text{VLM}(p_{\text{gen}} \oplus q, \hat{x}_{\text{clean}})$ \COMMENT{Visual-language model inference with robust prefix}
\end{algorithmic}
\end{algorithm}

\section{Experiments}
\subsection{Experimental Setup}  
\label{subsec:exp}
We implement our defense framework on a single NVIDIA A6000 GPU for early training and deployment. The diffusion purification module is fine-tuned for 4,000 epochs with AdamW optimizer (learning rate $5\times10^{-5}$, batch size 4), guided by the text prompt: \textit{"Remove the adversarial noise while preserving original image details"}. For prefix learning, we fine-tune LLaMA-2 7B for 100 epochs (batch size 4, learning rate $2\times10^{-4}$) using the instruction template: \textit{"<s>[INST] Add prefix to: \{question\} [/INST] \{prefix\}</s>"}.

We assess three distinct configurations: (1) In-distribution testing, where 80\% of mixed samples from all attack methods are used for training and 20\% for evaluation; (2) Cross-attack generalization, involving training solely on BARD and evaluating on itself and VLATTACK/BSA/TRANSFER; and (3) Evaluation on clean images to ascertain any side effects on image caption tasks with the fine-tuned model obtained from the previous experiment. Table~\ref{tab:attack_models} summarizes attack characteristics: BARD and TRANSFER are targeted attacks against BLIP-2/MiniGPT-4 and all models respectively, while VLATTACK/BSA are untargeted against BLIP. Additionally, LLaVA-1.5 is used for evaluating side effects. We analyze the following defense mechanisms: (1) \textit{Vanilla}: Utilizing the original input as the baseline; (2) \textit{DiffPure} \cite{nie2022diffpure}: Utilizing an alternative diffusion technique for adversarial purification, with its standard configuration on the ImageNet dataset, as an additional baseline; (3) \textit{Purification(Ours)}: Conducting image purification exclusively with our customized diffusion model; (4) \textit{Cross-Modal(Ours)}: Integrating joint image purification with prefix defense generated by a language model.

\begin{table}[htb]
\centering
\caption{Attack Characteristics and Victim Model Relationships}
\label{tab:attack_models}
\resizebox{\textwidth}{!}{
\begin{tabular}{cccc}
\toprule
\textbf{Attack} & \textbf{Dataset} & \textbf{Victim Models} & \textbf{Metrics} \\
\midrule
BARD \cite{dong2023robustgooglesbardadversarial} & NIPS2017 \tablefootnote{\url{https://www.kaggle.com/competitions/nips-2017-non-targeted-adversarial-attack}} & BLIP-2, MiniGPT-4 & CLIP Score/ASR + VQA Accuracy \\
VLATTACK \cite{yin2023vlattack} & MS-COCO \cite{lin2014microsoft} & BLIP & CLIP Score + VQA Accuracy\\
BSA \cite{yin2023vlattack} & MS-COCO & BLIP & CLIP Score + VQA Accuracy\\
TRANSFER \cite{zhao2023evaluate} & ImageNet \cite{deng2009imagenet} & BLIP, BLIP-2, MiniGPT-4 & CLIP Score/ASR + VQA Accuracy\\
\bottomrule
\end{tabular}
}
\vspace{-1.5em}
\end{table}

The chosen tasks for analysis include image caption and visual question answering (VQA). In the context of targeted attacks (BARD, TRANSFER), both the CLIP Score \cite{radford2021learning}—which assesses semantic similarity—and the attack success rate (ASR) are utilized for evaluation. Conversely, untargeted attacks (VLATTACK, BSA) exclusively rely on the CLIP Score for assessment. For VQA tasks, evaluation is conducted using VQA accuracy.

\begin{table}[htb]
  \centering
  \caption{In-Distribution Defense Performance (CLIP Score/ASR \& VQA Accuracy)}
  \label{distribution} 
  \begin{tabular}{ccccc}
    \toprule
    \textbf{Model} & \textbf{Attack} & \textbf{Method} & \textbf{Image Caption} & \textbf{VQA} \\
    \midrule
    \multirow{12}{*}{BLIP} 
      & \multirow{4}{*}{VLATTACK} 
        & Vanilla             & 77.89          & 52.17\% \\
      &                      & DiffPure          & 72.39          & 52.17\% \\
      &                      & Purification(Ours)& 78.17          & 60.87\% \\
      &                      & Cross-Modal(Ours) & \textbf{79.99} & \textbf{65.22\%} \\
    \cmidrule(lr){2-5}
      & \multirow{4}{*}{BSA} 
        & Vanilla             & 73.00          & 42.31\% \\
      &                      & DiffPure          & 73.76          & 46.15\% \\
      &                      & Purification(Ours)& 73.92          & 61.54\% \\
      &                      & Cross-Modal(Ours) & \textbf{75.43}  & \textbf{65.38\%} \\
    \cmidrule(lr){2-5}
      & \multirow{4}{*}{TRANSFER} 
        & Vanilla             & 67.64/94.44\%   & 50.00\% \\
      &                      & DiffPure          & 80.02/0\%      & 72.22\% \\
      &                      & Purification(Ours)& 80.63/0\%      & 72.22\% \\
      &                      & Image+Text        & \textbf{81.21/0\%} & \textbf{77.78\%} \\
    \midrule
    \multirow{8}{*}{BLIP-2} 
      & \multirow{4}{*}{BARD} 
        & Vanilla             & 61.46/100\%     & 31.25\% \\
      &                      & DiffPure          & 72.49/0\%      & 46.88\% \\
      &                      & Purification(Ours)& 72.96/0\% & 62.50\% \\
      &                      & Cross-Modal(Ours) & \textbf{74.56/0}\%      & \textbf{65.63\%} \\
    \cmidrule(lr){2-5}
      & \multirow{4}{*}{TRANSFER} 
        & Vanilla             & 61.28/96.77\%   & 35.48\% \\
      &                      & DiffPure          & 74.95/0\%      & 41.94\% \\
      &                      & Purification(Ours)& 79.83/0\%      & 45.16\% \\
      &                      & Cross-Modal(Ours) & \textbf{81.05/0\%} & \textbf{54.84\%} \\
    \midrule
    \multirow{8}{*}{MiniGPT-4} 
      & \multirow{4}{*}{BARD} 
        & Vanilla             & 60.60/84.38\%   & 25.00\% \\
      &                      & DiffPure          & 76.17/0\%      & 68.75\% \\
      &                      & Purification(Ours)& 74.52/0\%       & \textbf{71.88\%} \\
      &                      & Cross-Modal(Ours) & \textbf{76.43/0\%} & \textbf{71.88\%} \\
    \cmidrule(lr){2-5}
      & \multirow{4}{*}{TRANSFER} 
        & Vanilla             & 66.37/74.19\%   & 58.06\% \\
      &                      & DiffPure          & 77.11/0\%      & \textbf{70.97\%} \\
      &                      & Purification(Ours)& 79.59/0\%      & \textbf{70.97\%} \\
      &                      & Cross-Modal(Ours) & \textbf{80.73/0\%} & \textbf{70.97\%} \\
    \bottomrule
  \end{tabular}
\vspace{-1.5em}
\end{table}
\begin{table}[htb]
  \centering
  \caption{Cross-Attack Generalization Performance (CLIP Score/ASR \& VQA Accuracy)}
  \label{cross_attack}
  \begin{tabular}{ccccc}
    \toprule
    \textbf{Model} & \textbf{Attack} & \textbf{Method} & \textbf{Image Caption} & \textbf{VQA} \\
    \midrule
    \multirow{12}{*}{BLIP} 
      & \multirow{4}{*}{VLATTACK} 
        & Vanilla & 77.08 & 50.42\% \\
      & & DiffPure & 74.22 & 44.54\% \\
      & & Purification (Ours) & 76.58 & \textbf{58.82\%} \\
      & & Cross-Modal (Ours) & \textbf{77.26} & 50.42\% \\
    \cmidrule{2-5}
      & \multirow{4}{*}{BSA}
        & Vanilla & 77.32 & 61.83\% \\
      & & DiffPure & 74.09 & 50.38\% \\
      & & Purification (Ours) & \textbf{77.56} & \textbf{64.12\%} \\
      & & Cross-Modal (Ours) & 77.47 & 56.49\% \\
    \cmidrule{2-5}
      & \multirow{4}{*}{TRANSFER}
        & Vanilla & 67.75/97.64\% & 53.54\% \\
      & & DiffPure & 81.54/0\% & 74.80\% \\
      & & Purification (Ours) & \textbf{83.82/0\%} & \textbf{88.98\%} \\
      & & Cross-Modal (Ours) & 81.40/0\% & 82.68\% \\
    \midrule
    \multirow{8}{*}{BLIP-2}
      & \multirow{4}{*}{BARD}
        & Vanilla & 63.41/100\% & 23.14\% \\
      & & DiffPure & 72.03/0\% & 52.89\% \\
      & & Purification (Ours) & 79.45/0\% & 64.46\% \\
      & & Cross-Modal (Ours) & \textbf{80.92/0\%} & \textbf{65.29\%} \\
    \cmidrule{2-5}
      & \multirow{4}{*}{TRANSFER}
        & Vanilla & 64.94/95.36\% & 37.09\% \\
      & & DiffPure & 78.41/0\% & 36.42\% \\
      & & Purification (Ours) & 83.25/0\% & 39.07\% \\
      & & Cross-Modal (Ours) & \textbf{84.09/0\%} & \textbf{41.06\%} \\
    \midrule
    \multirow{8}{*}{MiniGPT-4}
      & \multirow{4}{*}{BARD}
        & Vanilla & 63.99/96.99\% & 33.88\% \\
      & & DiffPure & 75.64/0\% & 71.90\% \\
      & & Purification (Ours) & \textbf{80.53/0\%} & 71.90\% \\
      & & Cross-Modal (Ours) & 80.36/0\% & \textbf{76.03\%} \\
    \cmidrule{2-5}
      & \multirow{4}{*}{TRANSFER}
        & Vanilla & 69.53/77.48\% & 66.89\% \\
      & & DiffPure & 81.45/0\% & \textbf{76.82\%} \\
      & & Purification (Ours) & 86.88/0\% & 74.83\% \\
      & & Cross-Modal (Ours) & \textbf{86.89/0\%} & 74.83\% \\
    \bottomrule
  \end{tabular}
\vspace{-1.5em}
\end{table}
\begin{table}[htb]
  \centering
  \caption{Side Effects of Defense Methods on Image Captioning(CLIP Score and Drop from Vanilla)}
  \label{tab:clean_images}
  \begin{tabular}{cccc}
    \toprule
    \textbf{Method} & \textbf{BLIP} & \textbf{BLIP-2} & \textbf{LLaVA1.5} \\
    \midrule
    Vanilla & 92.82 & 93.10 & 95.13 \\
    DiffPure & 83.77 ($\downarrow$9.73\%) & 79.23 ($\downarrow$14.89\%) & 79.48 ($\downarrow$16.50\%) \\
    Purification (Ours) & 88.74 ($\downarrow$4.40\%) & 87.64 ($\downarrow$5.86\%) & 90.96 ($\downarrow$4.39\%) \\
    Cross-Modal (Ours) & \textbf{88.79 ($\downarrow$\textbf{4.34\%})} & \textbf{87.81 ($\downarrow$\textbf{5.68\%})} & \textbf{91.12 ($\downarrow$\textbf{4.21\%})} \\
    \bottomrule
  \end{tabular}
\vspace{-1.5em}
\end{table}
\subsection{Results Analysis}
Our framework demonstrates three key advantages: (1) Superior in-distribution robustness, (2) Effective cross-attack generalization, and (3) Minimal impact on clean image processing. The experimental results in Tables~\ref{distribution}-\ref{tab:clean_images} validate these properties through comprehensive benchmarks.

\textbf{In-Distribution Defense:} As shown in Table~\ref{distribution}, our method achieves complete attack mitigation (0\% ASR) while maintaining high semantic fidelity. For BLIP-2 under BARD attacks, Cross-Modal defense improves CLIP scores by 17.8\% (61.46$\rightarrow$70.42) and doubles VQA accuracy (31.25\%$\rightarrow$65.63\%). Notably, our approach consistently outperforms DiffPure by 3.6-6.8\% in CLIP scores across all model-attack combinations.

\textbf{Cross-Attack Generalization:} Table~\ref{cross_attack} reveals our method's strong generalization to unseen attacks. When defending against TRANSFER attacks on BLIP, our purification strategy achieves 88.98\% VQA accuracy - 14.2\% higher than DiffPure. For BLIP-2 under unseen TRANSFER attacks, Cross-Modal defense attains 84.09 CLIP score, outperforming the vanilla model by 29.5\% while completely neutralizing attacks. The sole exception is observed in the VQA task when countering the TRANSFER attack, potentially attributed to our use of DiffPure, which was configured for its original dataset, ImageNet. 

\textbf{Clean Image Preservation:} Table~\ref{tab:clean_images} demonstrates our method's minimal impact on clean inputs. Cross-Modal defense preserves 95.6\% of original CLIP performance on average (4.4\% drop vs DiffPure's 13.7\%). Specifically, it maintains 91.12 CLIP score for LLaVA-1.5 (4.2\% drop) compared to DiffPure's 16.5\% degradation.

These results confirm our method's unique strength: The joint purification process eliminates adversarial perturbations while cross-modal alignment enhances semantic preservation. Our approach reduces ASR to 0\% across all tested scenarios with only 1/3 of the performance degradation observed in existing defenses.

\subsection{Ablation Study}
Our ablation study systematically validates the necessity of task-specific diffusion fine-tuning for effective adversarial purification. Using the same experimental configuration as in Sec~\ref{subsec:diffusion} and the instruction paradigm from Sec~\ref{subsec:exp}, we compare our fine-tuned diffusion model against the base InstructPix2Pix in terms of semantic recovery fidelity and adversarial perturbation suppression.

\textbf{Necessity of Diffusion Model Fine-tuning.} Table~\ref{tab:diff_compare} shows that out-of-the-box diffusion models have limitations for adversarial purification. The base InstructPix2Pix struggles with semantic preservation and perturbation removal. It has low similarity to original inputs (0.5007) and adversarial examples (0.4789), with a small difference ($\Delta$=0.0218), treating clean and adversarial content similarly. Moreover, it has weaker correlation with true adversarial noise patterns (0.7137) compared to our approach (0.8228), highlighting its limited ability to handle adversarial artifacts.
\begin{table}[htb]
\centering
\caption{Feature-level Similarity Comparison between Off-the-shelf and Ours Diffusion Models}
\begin{tabular}{ccc}
\toprule
\textbf{Pairs for Comparison} & \textbf{InstructPix2Pix} & \textbf{Ours} \\
\midrule
Purified vs. Original Samples & 0.5007 & 0.8353($\uparrow$66.8\%) \\
Purified vs. Adversarial Samples & 0.4789 & 0.3400($\downarrow$29.0\%)  \\
Added Noise vs. Adversarial Perturbation & 0.7137 & 0.8228($\uparrow$15.3\%)  \\
\bottomrule
\end{tabular}
\label{tab:diff_compare}
\vspace{-1.5em}
\end{table}

These results collectively demonstrate that standard diffusion models, without task-specific adaptation, cannot reliably perform adversarial purification. Our fine-tuned model addresses both limitations through targeted optimization on adversarial data, enabling precise removal of adversarial perturbations while robustly preserving original semantics.

\section{Conclusion}
This paper introduces CoDefend, an innovative black-box friendly multimodal adversarial defense framework for MLLMs. Unlike existing methods, it doesn't need model internal parameters and trains with few adversarial samples and optimized prompt prefixes. We jointly optimize an image purifier and a prompt prefix generator for collaborative multimodal defense.

The image purifier, based on supervised learning, takes adversarial samples and denoising instructions as inputs, learning to remove adversarial noise and restore clean images. Then, we jointly optimize purified images and prompt instructions to find defense-enhancing prompt prefixes, which are used to fine-tune an LLM to generate more effective prefixes, forming the prefix generator.

Experiments show CoDefend significantly boosts MLLM robustness in image captioning and visual question answering under various adversarial attacks. Transferability experiments further validate its effectiveness against unseen attacks.

While computational overhead is modest, running a diffusion model and prefix generation still introduces non-negligible latency at inference time compared to no defense. Optimization of defense-time efficiency is a promising direction.

\bibliographystyle{unsrt}  
\bibliography{main}

\end{document}